\documentclass[runningheads]{llncs}
\usepackage[width=122mm,left=12mm,paperwidth=146mm,height=193mm,top=12mm,paperheight=217mm]{geometry}
\usepackage[utf8]{inputenc}
\DeclareMathAlphabet{\pazocal}{OMS}{zplm}{b}{n}
\usepackage{amssymb}
\usepackage{graphicx}
\usepackage{url}
\usepackage{subcaption}
\usepackage{epstopdf}
\usepackage{epsfig}
\usepackage{outlines}
\usepackage{multirow}
\usepackage{booktabs} 
\usepackage{amsmath}
\usepackage{mathrsfs}
\usepackage{paralist}
\usepackage{color}
\usepackage{float}
\usepackage{todonotes}
\usepackage[T1]{fontenc} 
\usepackage{url}
\usepackage{cite}
\usepackage{soul} 
\usepackage{multirow}
\usepackage{diagbox}
\usepackage{enumitem}
\usepackage{ulem}

\newcommand{\ie}{\textit{i}.\textit{e}., }

\newcommand{\tg}{\mathrm{TG}}

\newcommand{\gnn}{\mathrm{GNN}}
\newcommand{\cnn}{\mathrm{CNN}}

\newcommand{\concat}{\mathrm{Concat}}

\newcommand{\gin}{\mathrm{GIN}}
\newcommand{\wsss}{\mathrm{WSS}}

\newcommand{\mlp}{\mathrm{MLP}}
\newcommand{\prelu}{\mathrm{PReLU}}
\newcommand{\bn}{\mathrm{BN}}
\newcommand{\graphgradcam}{\textsc{GraphGrad-CAM}}
\newcommand{\gradcam}{\textsc{Grad-CAM}}
\newcommand{\gwsss}{\textsc{SegGini}}

\usepackage{hyperref}
\hypersetup{
    colorlinks   = true,
    citecolor    = green
}

\begin{document}

\mainmatter 
\title{Learning Whole-Slide Segmentation from Inexact and Incomplete Labels using Tissue Graphs}

\titlerunning{$\gwsss$}
\author{Valentin Anklin\inst{1,2\thanks{The authors contributed equally to this work.}}, Pushpak Pati\inst{1,2^\star}, Guillaume Jaume\inst{1,3^\star}, Behzad Bozorgtabar\inst{3}, Antonio Foncubierta-Rodríguez\inst{1}, Jean-Philippe Thiran\inst{3}, Mathilde Sibony\inst{4,5}, Maria Gabrani\inst{1}, Orcun Goksel\inst{2,6}}

\authorrunning{Anklin \textit{et\,al.}}

\institute{\textsuperscript{1} IBM Zurich Research Lab, Zurich, Switzerland \\
\textsuperscript{2} Computer-Assisted Applications in Medicine, ETH Zurich, Zurich, Switzerland\\
\textsuperscript{3} Signal Processing Laboratory 5, EPFL, Lausanne, Switzerland \\
\textsuperscript{4} APHP. Centre, Pathology department, Cochin hospital, Paris, France\\
\textsuperscript{5} Universit\'e de Paris, Paris, France\\
\textsuperscript{6} Department of Information Technology, Uppsala University, Sweden}

\maketitle

\begin{abstract}
Segmenting histology images into diagnostically relevant regions is imperative to support timely and reliable decisions by pathologists. To this end, computer-aided techniques have been proposed to delineate relevant regions in scanned histology slides.
However, the techniques necessitate task-specific large datasets of annotated pixels, which is tedious, time-consuming, expensive, and infeasible to acquire for many histology tasks.
Thus, weakly-supervised semantic segmentation techniques are proposed to utilize weak supervision that is cheaper and quicker to acquire.
In this paper, we propose $\gwsss$, a weakly supervised segmentation method using graphs, that can utilize weak \textit{multiplex} annotations, \ie  \textit{inexact} and \textit{incomplete} annotations, to segment arbitrary and large images, \textit{scaling} from tissue microarray (TMA) to whole slide image (WSI).
Formally, $\gwsss$ constructs a tissue-graph representation for an input histology image, where the graph nodes depict tissue regions. Then, it performs weakly-supervised segmentation via node classification by using \textit{inexact} image-level labels, \textit{incomplete} scribbles, or both.
We evaluated $\gwsss$ on two public prostate cancer datasets containing TMAs and WSIs. Our method achieved state-of-the-art segmentation performance on both datasets for various annotation settings while being comparable to a pathologist baseline.

\end{abstract}

\section{Introduction}
\label{introduction}

Automated delineation of diagnostically relevant regions in histology images is pivotal in developing automated computer-aided diagnosis systems in computational pathology. Accurate delineation assists the focus of the pathologists to improve diagnosis~\cite{wangs2019}. In particular, this attains high value in analyzing giga-pixel histology images. To this end, several supervised methods have been proposed to efficiently segment 
glands~\cite{sirinukunwattana2017,binder2019},
tumor regions~\cite{bejnordi2017,aresta2019}, and
tissue types \cite{bandi2017}.
Though these methods achieve high-quality semantic segmentation, they demand tissue, organ and task-specific dense pixel-annotated training datasets. However, acquiring such annotations for each diagnostic scenario is laborious, time-consuming, and often not feasible. Thus, weakly supervised semantic segmentation ($\wsss$) methods~\cite{zhou2017,chan2021}
are proposed to learn from weak supervision, such as \textit{inexact} coarse image labels, \textit{incomplete} supervision with partial annotations, and \textit{inaccurate} supervision where annotations may not always be ground-truth.

$\wsss$ methods employing various learning approaches, such as graphical model, multi-instance learning, self-supervised learning, are reviewed in~\cite{chan2021}. Further, $\wsss$ methods using various types of weak annotations are presented in~\cite{zhou2017,ahn2019}.
Despite the success in delivering excellent segmentation performance, mostly with natural images, $\wsss$ methods encounter challenges in histology images~\cite{chan2021}, since histology images contain,  
\begin{inparaenum}[(i)]
    \item finer-grained objects (\ie large intra- and inter-class variations)~\cite{xie2019}, and
    \item often ambiguous boundaries among tissue components~\cite{xu2017}.
\end{inparaenum}
Nevertheless, some $\wsss$ methods were proposed for histology.
Among those, the methods in~\cite{xu2014,hou2016,jia2017,xucamel2019,ho2021,lu2020} perform patch-wise image segmentation and cannot incorporate global tissue microenvironment context.
While \cite{chan2019,silva2021} propose to operate on larger image-tiles, they remain constrained to working with fixed and limited-size images.
Thus, a $\wsss$ method operating on arbitrary and large histology images by utilizing both local and global context is needed.
Further, most methods focus on binary classification tasks. Though HistoSegNet~\cite{chan2019} manages multiple classes, it requires training images with \textit{exact} fine-grained image-level annotations. \textit{Exact} annotations demand pathologists to annotate images beyond standard clinical needs and norms.
Thus, a $\wsss$ method should ideally be able to learn from \textit{inexact}, coarse, image-level annotations.
Additionally, to generalize to other $\wsss$ tasks in histology, methods should avoid complex, task-specific post-processing steps, as in HistoSegNet~\cite{chan2019}.
Notably, $\wsss$ methods in literature only utilize a single type of annotation. Indeed, complementary information from easily or readily available multiplex annotations can boost $\wsss$ performance. 

To this end, we propose $\gwsss$, a ``SEGmentation method using Graphs from Inexact and Incomplete labels''. 
$\gwsss$ constructs a superpixel-based tissue-graph representation for a histology image and follows a classification approach to segment the image.
Our major contributions are,
\begin{inparaenum}[(i)]
    \item $\gwsss$ is the first $\wsss$ method scalable to arbitrary image sizes, unlike pixel-based $\wsss$ or fully-connected graph-based $\wsss$~\cite{zhang2019,shi2020},
    \item to the best of our knowledge, $\gwsss$ is the first $\wsss$ method to simultaneously learn from weak multiplex supervision, \ie \textit{inexact} image-level labels as well as \textit{incomplete} scribbles.
    \item $\gwsss$ incorporates both local and global inter-tissue-region relations to build contextualized segmentation, principally in agreement with inter-pixel relation based state-of-the-art $\wsss$ method~\cite{ahn2019}.
\end{inparaenum}
 
\begin{figure}[t]
    \includegraphics[width=\linewidth]{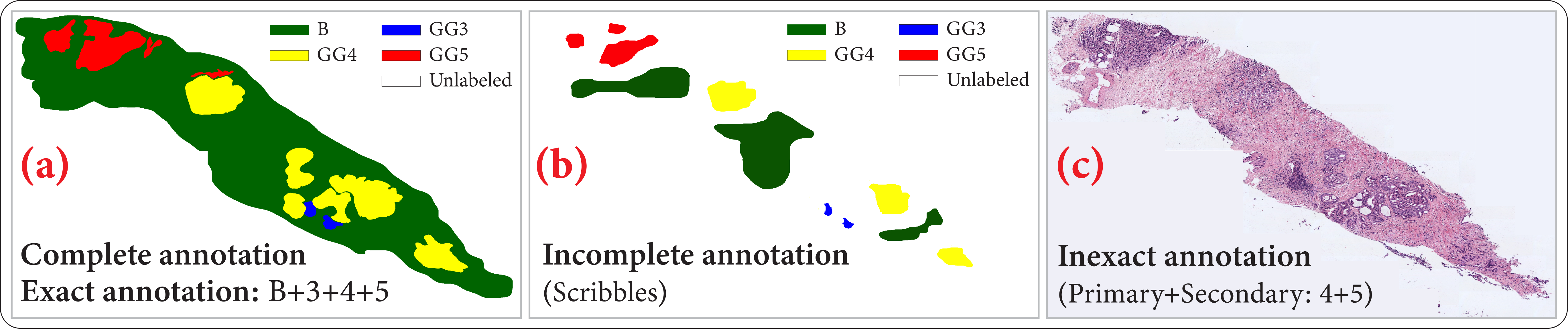}
    \caption{Overview of various annotation types for a sample prostate cancer WSI, (a) \textit{complete} pixel-level and \textit{exact} image-level annotation, (b) \textit{incomplete} scribbles of Gleason patterns, and (c) \textit{inexact} image-level Gleason grade (P+S).}
    \label{fig:annotations}
\end{figure}

We evaluate our method on two H\&E stained prostate cancer datasets~\cite{zhong2017,silva2020} containing TMAs and WSIs for segmenting Gleason patterns, \ie Benign (B), Grade3 (GG3), Grade4 (GG4) and Grade5 (GG5). 
To this end, we use \textit{incomplete} scribbles of Gleason patterns, and 
\textit{inexact} image-level Gleason grades. Image-level grades are defined the combination of the most common (\textit{primary}, P) and the second most common (\textit{secondary}, S) cancer growth patterns in the image.
Fig.~\ref{fig:annotations} exemplifies \textit{incomplete} and \textit{inexact} annotations, along with \textit{complete} pixel-level and \textit{exact} image-level annotation.

\vspace{-1em}
\section{Methods}
\begin{figure}[t]
    \includegraphics[width=\linewidth]{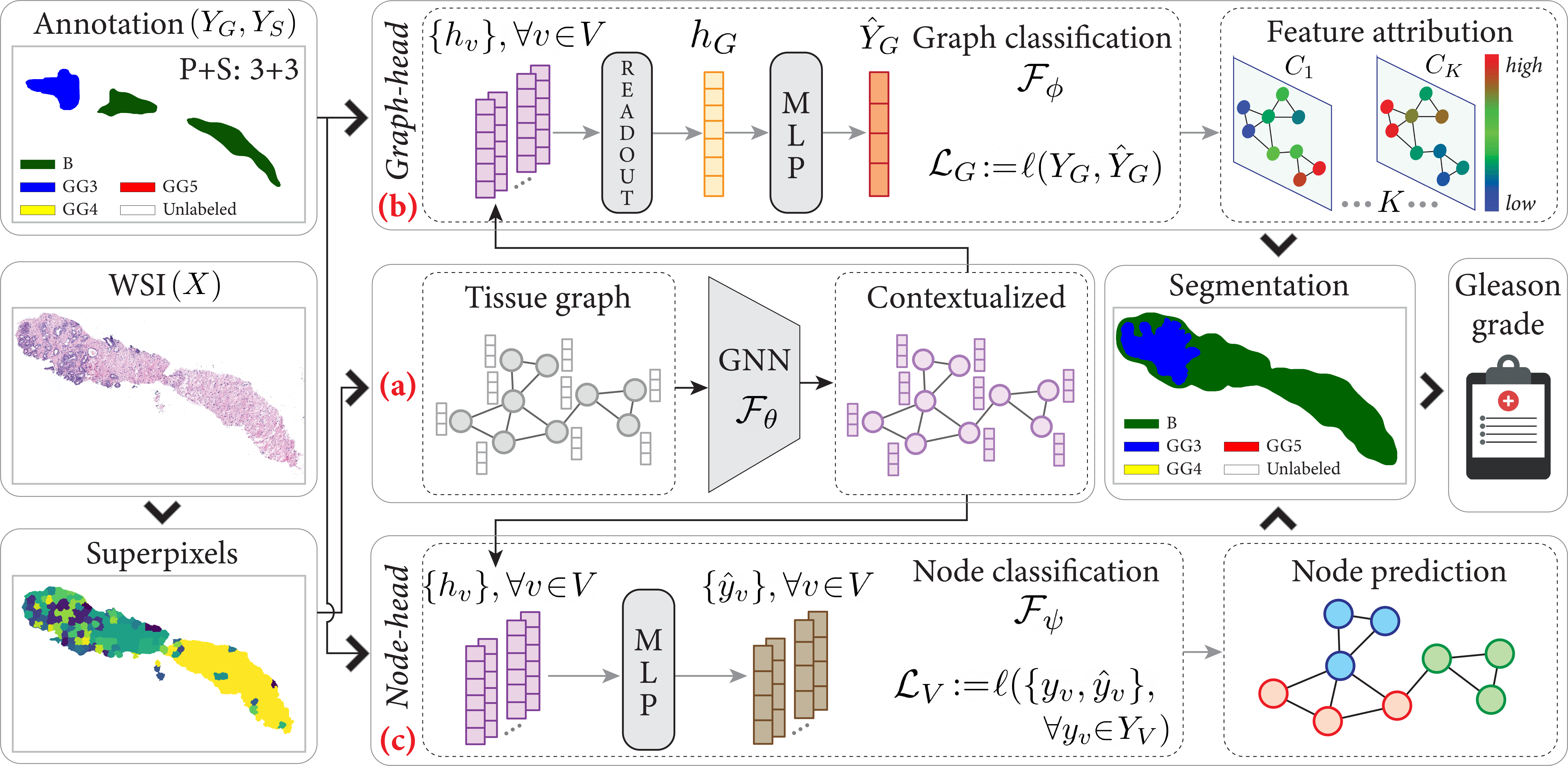}
    \caption{Overview of the proposed $\gwsss$ methodology. Following superpixel extraction, (a) Tissue-graph construction and contextualization, (b) \textit{Graph-head}: $\wsss$ via graph classification, (c) \textit{Node-head}: $\wsss$ via node classification. }
    \label{fig:block_diagram}
\end{figure}

This section presents the proposed $\gwsss$ methodology (Fig.~\ref{fig:block_diagram}) for scalable $\wsss$ of histology images. 
First, an input histology image is preprocessed and transformed into a tissue graph representation,
where the graph nodes denote tissue superpixels. Then, a Graph Neural Network ($\gnn$) learns contextualized features for the graph nodes.
The resulting node features are processed by a \textit{Graph-head}, a \textit{Node-head}, or both based on the type of weak supervision.
The outcomes of the heads are used to segment Gleason patterns. Additionally, a classification is performed to identify image-level Gleason grades from the segmentation map.

\vspace{1.5ex} \noindent \textbf{Preprocessing and Tissue Graph Construction. }
An input H\&E stained histology image $X$ is stain-normalized to reduce any appearance variability due to tissue preparation using the unsupervised stain normalization algorithm in~\cite{vahadane2016}.
Then, the normalized image is transformed into a Tissue-Graph ($\tg$) (Fig.~\ref{fig:block_diagram}\textcolor{red}{(a)}), as proposed in~\cite{pati2020}. Formally, we define a $\tg$ as $G := (V, E, H)$, where the nodes $V$ encode meaningful tissue regions in the form of \textit{superpixels}, and the edges $E$ represent inter-tissue interactions.
Each node $v \in V$ is represented by a feature vector $h(v) \in \mathbb{R}^d$. We denote the node features set, $h(v), \, \forall v \in V$ as $H \in \mathbb{R}^{|V| \times d}$. Motivated by~\cite{bejnordi2015}, we use superpixels as visual primitives, since rectangular patches may span multiple distinct structures. 

The $\tg$ construction follows three steps:
\begin{inparaenum}[(i)]
\item superpixel construction to define $V$,
\item superpixel feature extraction to define $H$, and
\item graph topology construction to define $E$.
\end{inparaenum}  
For superpixels, we first use the unsupervised SLIC algorithm~\cite{achanta2011} emphasizing on space proximity. Over-segmented superpixels are produced at a lower magnification to capture homogeneity, offering a good compromise between granularity and noise smoothing.
The superpixels are hierarchically merged based on channel-wise color similarity of superpixels at higher magnification, \ie channel-wise 8-bin color histograms, mean, standard-deviation, median, energy, and skewness. These then form the $\tg$ nodes. 
The merging reduces node complexity in the $\tg$, thereby enabling a scaling to large images and contextualization to distant nodes, as explained in next section.
To characterize the $\tg$ nodes, we extract morphological and spatial features. Patches of 224$\times$224 are extracted from the original image and encoded into 1280-dimensional features with MobileNetV2~\cite{sandler2018} pre-trained on ImageNet~\cite{deng2009}.
For a node $v \in V$, morphological features are computed as the mean of individual patch-level representations that belong to $v$. 
Spatial features are computed by normalizing superpixel centroids by the image size. 
We define the $\tg$ topology by constructing a region adjacency graph (RAG)~\cite{potjer1996} from the spatial connectivity of superpixels.

\vspace{1.5ex}\noindent\textbf{Contextualized Node Embeddings. }
\label{sec:node_embedding}
Given a $\tg$, we aspire to learn discriminative node embeddings (Fig.~\ref{fig:block_diagram}\textcolor{red}{(a)}) that benefit from the nodes' context, \ie the tissue microenvironment and inter-tissue interactions. The contextualized node embeddings are further used to perform semantic segmentation.
To this end, we employ a $\gnn$, a family of networks able to operate on graph-structured data~\cite{kipf2017,xu2019,dwivedi2020}.
In particular, we use Graph Isomorphism Network ($\gin$)~\cite{xu2019} layers, a powerful and fast $\gnn$ architecture that functions as follows.
For each node $v \in V$, $\gin$ uses a \textit{sum}-operator to \textit{aggregate} the features of the node's neighbors $\mathcal{N}(v)$.
Then, it \textit{updates} the node features $h(v)$ by combining the \textit{aggregated} features with the current node features $h(v)$ via a multi-layer perceptron ($\mlp$).
After $T$ $\gin$ layers, \ie acquiring context up to $T$-hops, the intermediate node features $h^{(t)}(v), \; t=1,\dots, T$ are concatenated to define the contextualized node embeddings~\cite{xu2018}.
Formally, a $\gnn$ $\mathcal{F}_\theta$ with batch normalization (BN) is described for $v, u \in V$ as,
\begin{align} 
\label{eq:node_update}
    h^{(t+1)}(v) &= \mlp\Big( \bn \big( h^{(t)}(v) + \sum_{u \in \mathcal{N}(v)} h^{(t)}(u) \big) \Big), \; t=\{0, .., T-1\}\,
    \\
    h(v) &= \concat\,\Big(\,\Big\{\,h^{(t)}(v) \;\Big|\;t = 1, .., T \, \Big\} \, \Big)
\end{align}

\vspace{.5ex}\noindent\textbf{Weakly Supervised Semantic Segmentation. }
The contextualized node embeddings $h(v), \, \forall v \in V$ for a graph $G$, corresponding to an image $X$, are processed by $\gwsss$ to assign a class label $\in \{1,..,K\}$ to each node $v$, where $K$ is the number of semantic classes. $\gwsss$ can incorporate multiplex annotations,
\ie \textit{inexact} image label $Y_X$ and \textit{incomplete} scribbles $Y_S$.
Then, the weak supervisions for $G$ are, the graph label $Y_G$, \ie the image label $Y_X$, and node labels $y_v \in Y_V$ that are extracted from $Y_S$ by assigning the most prevalent class within each node.
This is a reasonable assumption, as the tissue regions are built to be semantically homogeneous.
The \textit{Graph-head} (Fig.~\ref{fig:block_diagram}\textcolor{red}{(b)}) and the \textit{Node-head} (Fig.~\ref{fig:block_diagram}\textcolor{red}{(c)}) are executed for using $Y_G$ and $Y_V$, respectively.
Noticeably, unlike~\cite{chan2019}, $\gwsss$ does not involve any post-processing, thus being a generic method that can be applied to various organs, tissue types, segmentation tasks, etc.

The \textit{Graph-head}
consists of a graph classification and a feature attribution module.
First, a graph classifier $\mathcal{F}_\phi$ predicts $\hat{Y}_G$ for $G$. $\mathcal{F}_\phi$ includes, (i) a global average pooling \textit{readout} operation to produce a fixed-size graph embedding $h_G$ from the node embeddings $h(v),\forall v \in V$, and (ii) a $\mlp$ to map $h_G$ to $Y_G$. As $G$ directly encodes $X$, the need for patch-based processing is nullified.
$\mathcal{F}_\theta$ and $\mathcal{F}_\phi$ are trained on a graph-set $\mathcal{G}$, extracted from the image-set $\mathcal{X}$, by optimizing a multi-label weighted binary cross-entropy loss $\mathcal{L}_G:= l(Y_G, \hat{Y}_G)$.
The class-weights are defined by $w_i = \log(N/n_i), i=1,...,K$, where $N=|\mathcal{X}|$, and $n_i$ is the class example count; such that higher weight is assigned to smaller classes to mitigate class imbalance during training.
Second, in an off-line step, we employ a discriminative \textit{feature attribution} technique to measure importance scores $\forall v \in V$ towards the classification of each class. Specifically, we use $\graphgradcam$~\cite{pope2019, jaume2021}, a version of $\gradcam$~\cite{selvaraju2017} that can operate with $\gnn$s. \textit{Argmax} across class-wise node attribution maps from $\graphgradcam$ determines the node labels.

The \textit{Node-head}
simplifies image segmentation into classifying nodes $v \in V$. It inputs $h(v),\,\forall v \in V$ to a $\mlp$ classifier $\mathcal{F}_\psi$ to predict node-labels $y_v,\,\forall v \in V$.
$\mathcal{F}_\theta$ and $\mathcal{F}_\psi$ are trained using the multi-class weighted cross-entropy loss $\mathcal{L}_V:= l(y_v, \hat{y}_v)$.
The class-weights are defined by $w_i = \log(N/n_i), i=1,...,K$, where $N$ is the number of annotated nodes, and $n_i$ is the class node count.
The node-wise predicted classes produce the final segmentation.

\textit{Multiplexed supervision:} 
For multiplex annotations, both heads are executed to perform $\wsss$. $\mathcal{F}_\theta$, $\mathcal{F}_\phi$, and $\mathcal{F}_\psi$ are jointly trained to optimize a weighted loss $\mathcal{L} = \lambda \mathcal{L}_G + (1 - \lambda) \mathcal{L}_V$, with which complementary information from multiplex annotations helps improve the individual classification tasks and thus improving $\wsss$.
Subsequently, we employ the classification approach in \cite{arvaniti2018} to determine the Gleason grades from the generated segmentation maps.

\section{Experiments}
\label{sec:experiments}

We evaluate our method on 2 prostate cancer datasets for Gleason pattern segmentation and Gleason grade classification.

\vspace{1ex} \noindent\textbf{UZH dataset}~\cite{zhong2017} comprises five TMAs with 886 spots, digitized at 40$\times$ resolution (0.23\,$\mu$m/pixel). Spots (3100$\times$3100 pixels) contain \textit{complete} pixel-level annotations and \textit{inexact} image-level grades.
We follow a 4-fold cross-validation at TMA-level with testing on TMA-80 as in \cite{arvaniti2018}.
The second pathologist annotations on the test TMAs are used as a pathologist-baseline.

\vspace{1ex} \noindent\textbf{SICAPv2 dataset}~\cite{silva2020} contains 18\,783 patches of size 512$\times$512 with \textit{complete} pixel annotations and WSI-level grades from 155 WSIs at 10$\times$ resolution.
We reconstruct the original WSIs and annotation masks from the patches,  containing up to $11000^2$ pixels. 
We follow a 4-fold cross-validation at patient-level as in \cite{silva2020}. An independent pathologist's annotations are included as a pathologist-baseline.

\begin{figure}[t]
    \centering
    \includegraphics[width=0.95\linewidth]{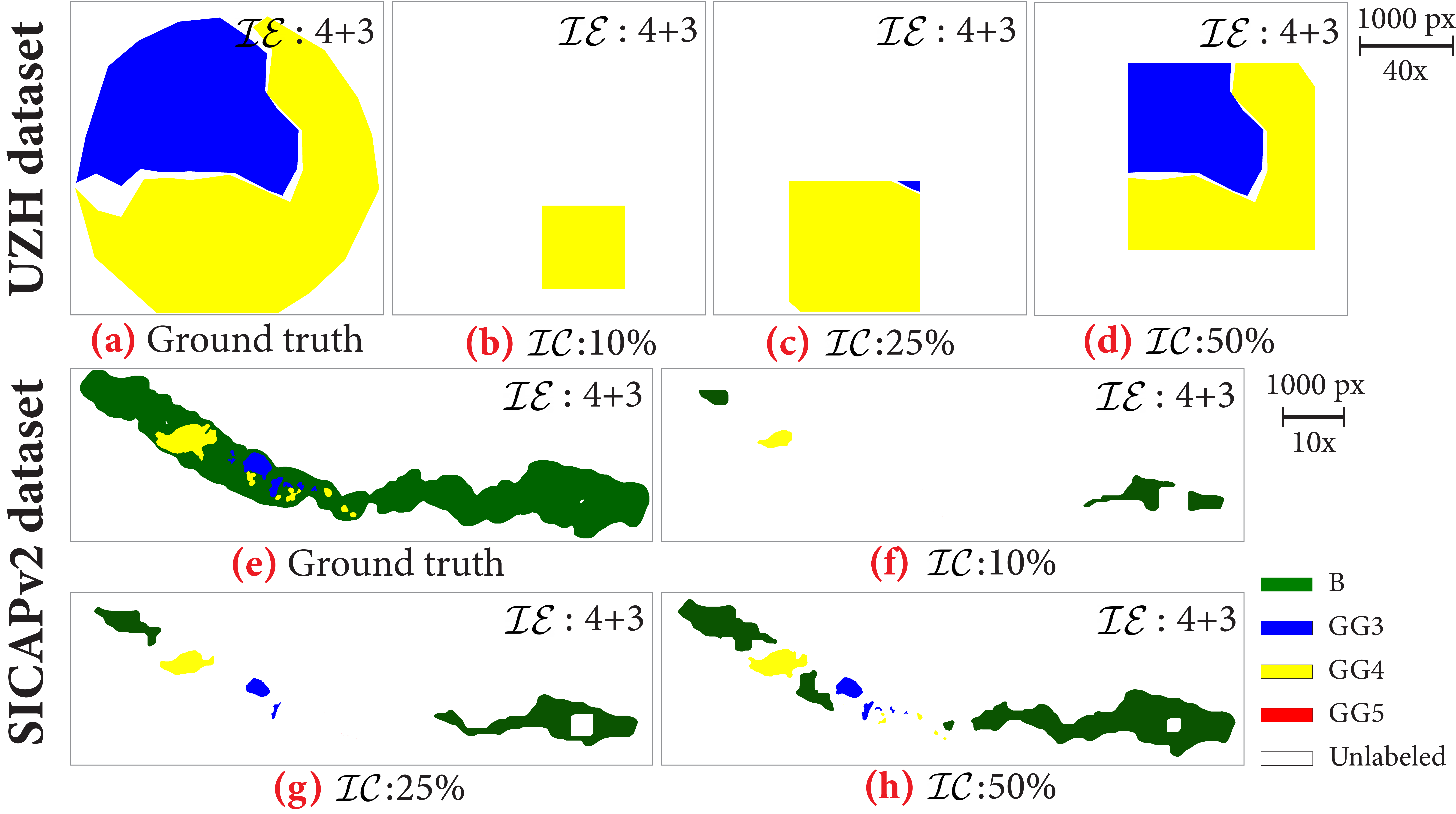}
    \caption{Example of $\mathcal{IE}$ and $\mathcal{IC}$ annotations for various percentage of $\mathcal{IC}$.}
    \label{fig:annotationPercent}
\end{figure}

We evaluate the methods for four annotation settings, \textit{complete} ($\mathcal{C}$) and \textit{incomplete} ($\mathcal{IC}$) pixel annotations, \textit{inexact} image labels ($\mathcal{IE}$) as well as $\mathcal{IE+IC}$.
$\mathcal{IC}$ annotations with various pixel percentages are created by randomly selecting regions from $\mathcal{C}$, as shown in Fig.~\ref{fig:annotationPercent}.
We report per-class and average Dice scores as segmentation metrics, and  weighted F1-score as a classification metric.
We present means and standard-deviations on the test set for 4-fold cross-validation for all experiments. The fold-wise dataset statistics are presented in Tab.~\ref{tab:stats}. 

\vspace{-2em}
\begin{table}[hbt!]
\caption{Fold-wise dataset statistics of Gleason patterns in train (Tr), val (V) and test (Te).}
\label{tab:stats}
\centering
\begin{tabular}{lc|c|c|c}
\toprule
\multicolumn{5}{c}{UZH}  \\
\toprule
Fold$\sim$Tr/V/Te & Benign & Grade3 & Grade4 & Grade5 \\
\midrule
Fold 1 & $273/121/78$ & $336/82/196$ & $276/46/202$ & $165/23/25$ \\
Fold 2 & $392/2/78$ & $414/4/196$ & $255/67/202$ & $51/137/25$ \\
Fold 3 & $260/134/78$ & $271/147/196$ & $256/66/202$ & $182/6/25$ \\
Fold 4 & $257/137/78$ & $233/185/196$ & $179/143/202$ & $166/22/25$ \\
\midrule

\multicolumn{5}{c}{SICAPv2}  \\
\toprule
Fold$\sim$Tr/V/Te & Benign & Grade3 & Grade4 & Grade5 \\
\midrule
Fold 1 & $50/10/12$ & $56/16/20$ & $64/26/25$ & $20/6/5$ \\
Fold 2 & $48/12/12$ & $64/8/20$ & $62/28/25$ & $20/6/5$ \\
Fold 3 & $42/18/12$ & $46/26/20$ & $78/12/25$ & $22/4/5$ \\
Fold 4 & $40/20/12$ & $50/22/20$ & $66/24/25$ & $16/10/5$ \\
\bottomrule
\end{tabular}
\end{table}

\vspace{1ex} \noindent \textbf{Baselines:}
We compare $\gwsss$ with several state-of-the-art methods:
\begin{itemize}[nosep]
    \item UZH-$\cnn$~\cite{arvaniti2018} and FSConv~\cite{silva2020}, for segmentation and classification using $\mathcal{C}$
    \item Neural Image Compression (NIC)~\cite{tellez2021}, Context-Aware $\cnn$ (CACNN)~\cite{shaban2020}, and CLAM~\cite{lu2020}, for weakly-supervised classification using $\mathcal{IE}$
    \item HistoSegNet~\cite{chan2019}, for weakly supervised segmentation using $\mathcal{IE}$.
\end{itemize}
These baselines are implemented based on code and algorithms in the corresponding publications. 
Baselines~\cite{lu2020,tellez2021,shaban2020} directly classify WSI Gleason grades, and do not provide segmentation of Gleason patterns.
Also, HistoSegNet~\cite{chan2019} was trained herein with $\mathcal{IE}$, instead of \textit{exact} image labels, since accessing the \textit{exact} annotations would require using $\mathcal{C}$, that violates weak supervision constraints.

\vspace{1ex} \noindent \textbf{Training and Implementation} 
were conducted using PyTorch~\cite{paszke2019} and DGL library~\cite{wang2019} on an NVIDIA Tesla P100.
$\gwsss$ model consists of 6-$\gin$ layers, where the $\mlp$ in $\gin$, the \textit{graph-head}, and the \textit{node-head} contain 2-layers each with $\prelu$ activation and 32-dimensional node embeddings, inspired by~\cite{You2020}.
For graph augmentation, the superpixel nodes were augmented randomly with rotation and mirroring.
A hyper-parameter search was conducted to find the optimal batch size $\in \{4, 8, 16\}$, learning rate $\in \{10^{-3}, 5 \times 10^{-4}, 10^{-4} \}$, dropout $\in \{ .25, .5\}$, and $\lambda \in \{.25, .5, .75\}$ for each setting. The methods were trained with Adam optimizer to select the model with best validation Dice score.
To ensure consistent and comparable comparisons, we evaluated all the baselines with similar patch-level augmentations and hyper-parameter searches.

\newcommand{\std}[1]{{\scriptsize$\pm$#1}}
\begin{table}[t]
\caption{Results on UZH dataset as \textit{Mean\std{std}} using complete ($\mathcal{C}$), inexact ($\mathcal{IE}$), incomplete ($\mathcal{IC}$), and $\mathcal{IE+IC}$ settings. Setting-wise best scores are in \textbf{bold}.}
\label{tab:uzh_results}
\centering
\scalebox{1}{
\begin{tabular}{l|l@{~~}l@{~~}l@{~~}l@{~~}l@{~~}l@{~~}l}
\toprule
\parbox[t]{3mm}{\multirow{2}{*}[0ex]{\rotatebox[origin=c]{90}{Annot.}}} & & \multicolumn{4}{c}{per-class Dice}  & avg. Dice & weight-F1         \\
\cmidrule(lr){3-6}
& Method  & Benign & Grade3 & Grade4 & Grade5 & & \\
\midrule
\parbox[t]{3mm}{\multirow{2}{*}[0ex]{\rotatebox[origin=c]{90}{$\mathcal{C}$}}} & UZH-$\cnn$\cite{arvaniti2018} & \textbf{69.5\std{6.0}} & 54.7\std{3.9} & 63.6\std{3.2} & 34.6\std{4.6} & 55.6\std{1.8} & 49.2\std{4.3} \\
& $\gwsss$ & 64.2\std{8.0} & \textbf{71.3\std{1.9}} & \textbf{72.9\std{2.8}} & \textbf{55.6\std{3.3}} & \textbf{66.0\std{3.1}} & \textbf{56.8\std{1.7}} \\
\midrule
\parbox[t]{3mm}{\multirow{5}{*}[-0.3ex]{\rotatebox[origin=c]{90}{$\mathcal{IE}$}}} & CLAM\cite{lu2020} & - & - & - & - & - & 45.7\std{4.6} \\
& NIC\cite{tellez2021} & - & - & - & - & - & 33.5\std{5.5} \\
& CACNN\cite{shaban2020} & - & - & - & - & - & 26.1\std{5.1} \\
& HistoSegNet\cite{chan2019} & \textbf{89.0\std{3.8}} & 42.4\std{10.9} & 56.8\std{10.4} & 34.8\std{12.9} & 55.7\std{3.2} & 41.6\std{9.3}\\
& $\gwsss$ & 63.0\std{9.3} & \textbf{69.6\std{5.6}} & \textbf{67.6\std{5.4}} & \textbf{55.7\std{7.0}} & \textbf{64.0\std{1.8}} & \textbf{52.4\std{3.2}} \\
\midrule
\multicolumn{1}{c}{} & Pathologist & 83.33 & 44.53 & 69.29 & 57.28 & 63.60 & 48.98\\
\bottomrule
\end{tabular}
}

\scalebox{1}{
\begin{tabular}{l@{\hspace{0.5cm}}|c@{\hspace{0.5cm}}l@{\hspace{0.5cm}}l@{\hspace{0.5cm}}l@{\hspace{0.5cm}}l}
\multicolumn{2}{l}{}    & \multicolumn{4}{c}{avg. Dice}  \\
\cmidrule(lr){3-6}
Annot. & Method & 5\% pixel & 10\% pixel & 25\% pixel & 50\% pixel \\
\midrule
$\mathcal{IC}$ & $\gwsss$ & 58.2\std{3.1} & 62.9\std{2.2} & 63.3\std{2.3} & \textbf{65.3\std{3.1}} \\   
$\mathcal{IC+IE}$ & $\gwsss$ & \textbf{63.7\std{2.9}} & \textbf{65.6\std{3.0}} & \textbf{63.6\std{2.0}} & 64.2\std{2.6} \\
\bottomrule
\end{tabular}}
\end{table}

\vspace{1ex} \noindent \textbf{Results and Discussion:}
Tab.~\ref{tab:uzh_results} and \ref{tab:sicapv2_results} present the segmentation and classification results of $\gwsss$ and the baselines, divided in groups for their use of different annotations.
For the $\mathcal{C}$ setting, $\gwsss$ significantly outperforms UZH-CNN~\cite{arvaniti2018} on per-class and average segmentation as well as classification metrics, while reaching segmentation performance comparable with pathologists. 
For the $\mathcal{IE}$ setting, $\gwsss$ outperforms HistoSegNet on segmentation and classification tasks. 
Interestingly, $\gwsss$ also outperforms the classification-tailored baselines~\cite{lu2020,tellez2021,shaban2020}. 
$\gwsss$ delivers comparable segmentation performance for \textit{inexact} and \textit{complete} supervision, \ie 64\% and 66\% average Dice, respectively. 
Comparing $\mathcal{IC}$ and $\mathcal{IE+IC}$, we observe that $\mathcal{IE+IC}$ produces better segmentation, especially in the low pixel-annotation regime. Such improvement, however, lessens with increased pixel annotations, which is likely due to the homogeneous Gleason patterns in the test set with only one or two patterns per TMA. 
Notably, $\gwsss$ with $\mathcal{IE}$ setting outperforms UZH-$\cnn$ with $\mathcal{C}$ setting.

On SICAPv2 dataset in $\mathcal{C}$ setting, $\gwsss$ outperforms FSConv on both segmentation and classification tasks, and performs comparable to the pathologist-baseline for classification.
SICAPv2 is a highly imbalanced dataset with a large fraction of benign regions. Thus, $\gwsss$ yields better results for benign class, while relatively poor performance for Grade5, which is rare in the dataset. 
For the $\mathcal{IE}$ setting, $\gwsss$ significantly outperforms HistoSegNet that trains using tile-labels, set the same as WSI-labels. This indicates that HistoSegNet is not applicable to WSIs with WSI-level supervision. 
For $\mathcal{IE}$, $\gwsss$ performs superior to \cite{tellez2021,shaban2020} and comparable to \cite{lu2020}.
Combining $\mathcal{IE}$ and $\mathcal{IC}$ for segmentation, the complementarity of annotations substantially boosts $\gwsss$ performance. 
$\gwsss$ with $\mathcal{IE+IC}$ setting consistently outperforms $\mathcal{IC}$ for various \% of pixel annotations.
Notably, $\mathcal{IE+IC}$ outperforms $\mathcal{C}$ while using only 50\% pixels. This confirms the benefit of learning from \textit{multiplex} annotations. 

\begin{figure}[h]
    \centering
    \includegraphics[width=0.85\linewidth]{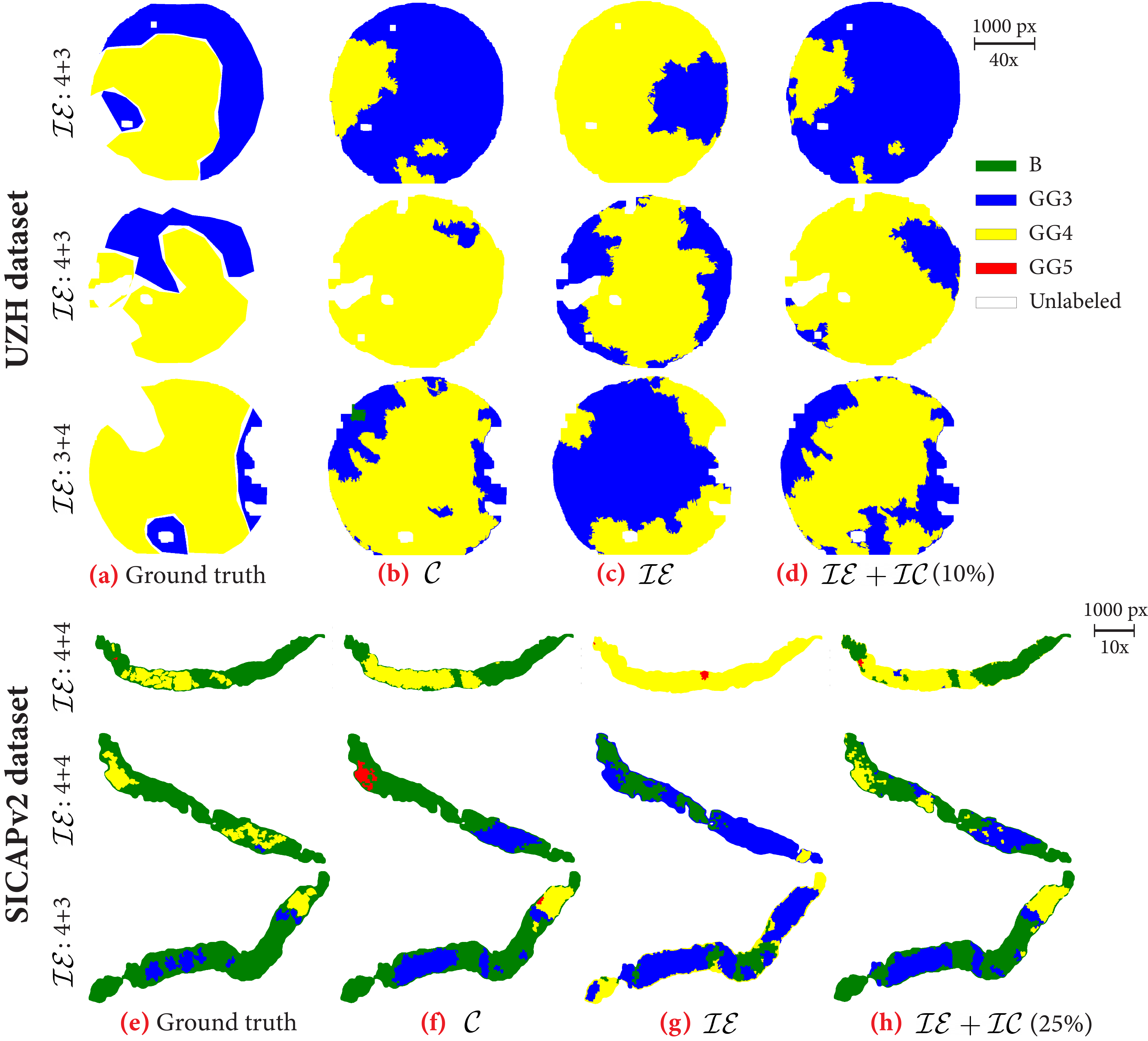}
    \caption{Predicted segmentation maps for various annotation settings, \ie \textit{complete} ($\mathcal{C}$), \textit{inexact} ($\mathcal{IE}$),  \textit{inexact + incomplete} ($\mathcal{IE+IC}$).}
    \label{fig:qualititative_results}
\end{figure}

\begin{figure}[h]
    \centering
    \includegraphics[width=0.85\linewidth]{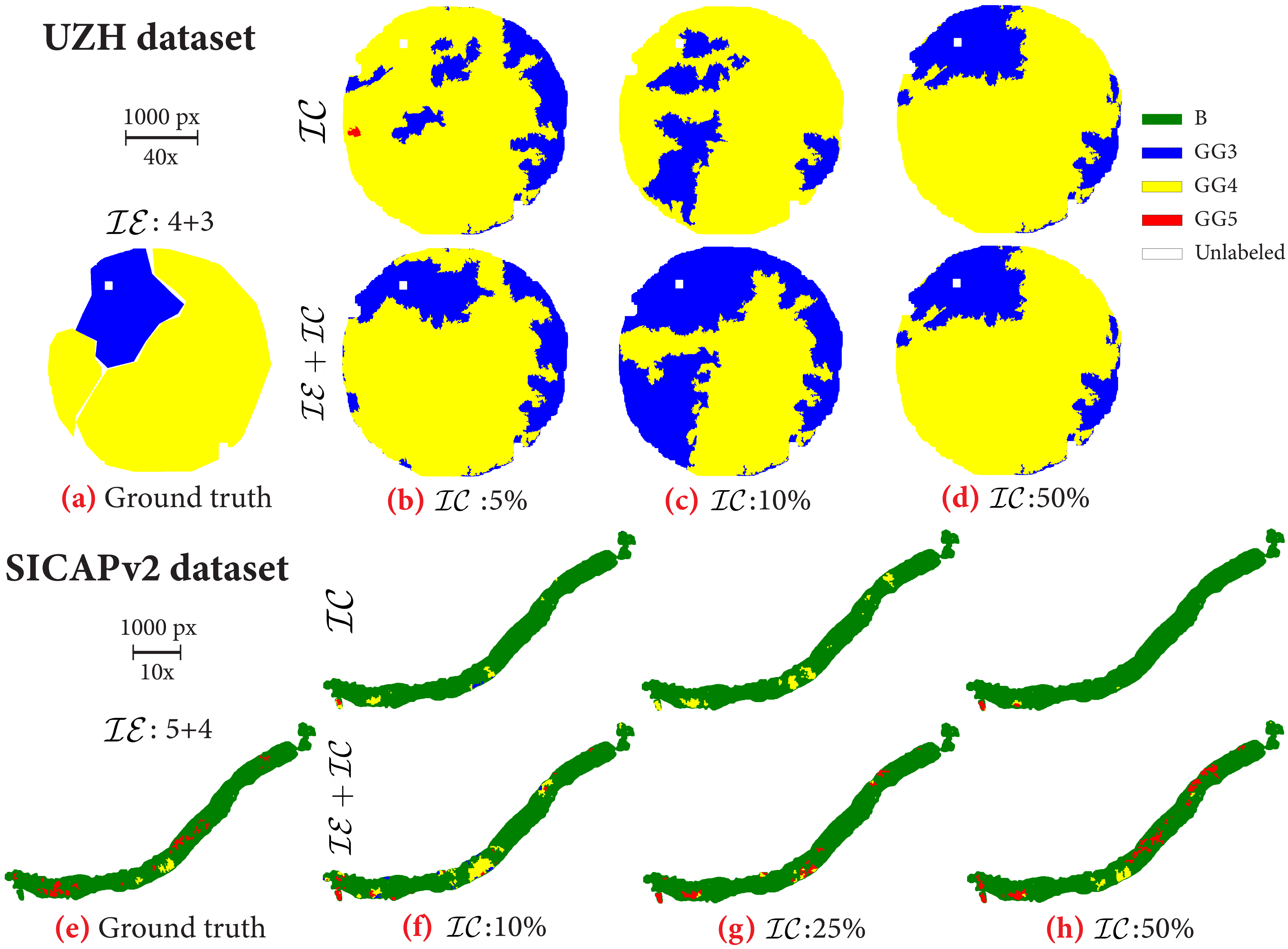}
    \caption{Example of $\mathcal{IE}$ and $\mathcal{IC}$ annotations for various percent of $\mathcal{IC}$.}
    \label{fig:ieic_percentages}
\end{figure}

Fig.~\ref{fig:qualititative_results} presents qualitative results on both datasets for various annotation settings. 
Fig.~\ref{fig:ieic_percentages} presents qualitative results on both datasets for $\mathcal{IC}$ and $\mathcal{IE+IC}$ settings with various percentages of $\mathcal{IC}$.
$\mathcal{IE+IC}$ produces satisfactory segmentation while correcting any errors in $\mathcal{IE}$ by incorporating scribbles.
The results indicate that $\gwsss$ provides competitive segmentation even with \textit{inexact} supervision. Thus, we can leverage readily available slide-level Gleason grades from clinical reports, to substantially boost the segmentation, potentially together with a few \textit{incomplete} scribbles from pathologists.

\begin{table}[t]
\caption{Results on SICAPv2  as \textit{Mean\std{std}} using complete ($\mathcal{C}$), inexact ($\mathcal{IE}$), incomplete ($\mathcal{IC}$), and $\mathcal{IE+IC}$ settings. Setting-wise best scores are in \textbf{bold}.}
\label{tab:sicapv2_results}
\centering

\scalebox{0.96}{
\begin{tabular}{l|l@{~~}l@{~~}l@{~~}l@{~~}l@{~~}l@{~~}l}
\toprule
\parbox[t]{3mm}{\multirow{2}{*}[0ex]{\rotatebox[origin=c]{90}{Annot.}}} & & \multicolumn{4}{c}{per-class Dice}  & avg. Dice & weight-F1         \\
\cmidrule(lr){3-6}
 & Method & Benign & Grade3 & Grade4 & Grade5 & &  \\
\midrule
\parbox[t]{3mm}{\multirow{2}{*}[0ex]{\rotatebox[origin=c]{90}{$\mathcal{C}$}}}
& FSConv~\cite{silva2020} & 59.4\std{3.0} & 23.7\std{2.6} & 30.7\std{2.7} & \textbf{9.1\std{2.9}} & 31.3\std{2.5} & 59.9\std{5.0} \\
& $\gwsss$ & \textbf{90.0\std{0.1}} & \textbf{39.4\std{3.3}} & \textbf{40.2\std{2.7}} & 7.4\std{2.4} & \textbf{44.3\std{2.0}} & \textbf{62.0\std{3.6}} \\
\midrule

\parbox[t]{3mm}{\multirow{5}{*}[-0.3ex]{\rotatebox[origin=c]{90}{\shortstack[c]{$\mathcal{IE}$}}}}
& CLAM\cite{lu2020} & - & - & - & - & - & 47.5\std{4.3} \\
& NIC\cite{tellez2021} & - & - & - & - & - & 32.4\std{10.0} \\
& CACNN\cite{shaban2020} & - & - & - & - & - & 21.8\std{4.7} \\
& HistoSegNet\cite{chan2019} & \textbf{78.1\std{1.4}} & 1.5\std{0.7} & 8.4\std{0.9} & 1.6\std{0.3} & 22.4\std{0.3} & 16.7\std{4.3} \\
& $\gwsss$ & 55.9\std{12.0} & \textbf{19.5\std{6.7}} & \textbf{20.7\std{2.9}} & \textbf{8.0\std{4.2}} & \textbf{26.0\std{5.0}} & \textbf{48.7\std{6.3}} \\
\midrule 
\multicolumn{1}{c}{} & Pathologist & - & - & - & - & - & 63.00 \\
\bottomrule
\end{tabular}
}

\scalebox{0.96}{
\begin{tabular}{l@{\hspace{0.5cm}}|l@{\hspace{0.5cm}}l@{\hspace{0.5cm}}l@{\hspace{0.5cm}}l@{\hspace{0.5cm}}l}
\multicolumn{2}{l}{}    & \multicolumn{4}{c}{avg. Dice}  \\
\cmidrule(lr){3-6}
Annot. & Method & 10\% pixel & 25\% pixel & 50\% pixel & 100\% pixel\\
\midrule
$\mathcal{IC}$ & $\gwsss$ & 37.8\std{1.1} & \textbf{41.9\std{1.0}} & 42.4\std{0.8} & 44.3\std{2.0} \\
$\mathcal{IC+IE}$ & $\gwsss$ & \textbf{39.6\std{1.2}} & 41.8\std{0.6} & \textbf{46.0\std{0.6}} & \textbf{47.0\std{1.8}} \\
\bottomrule
\end{tabular}}
\end{table}

\section{Conclusion}
\label{conclusion}
We proposed a novel $\wsss$ method, $\gwsss$, to perform semantic segmentation of histology images by leveraging complementary information from weak multiplex supervision, \ie \textit{inexact} image labels and \textit{incomplete} scribbles. $\gwsss$ employs a graph-based classification that can directly operate on large histology images, thus utilizing local and global context for improved segmentation. $\gwsss$ is a generic method that can be applied to different tissues, organs, and histology tasks. We demonstrated state-of-the-art segmentation performance on two prostate cancer datasets for various annotation settings, while not compromising on classification results. Future research will focus on studying the generalizability of our method to previously unseen datasets.
    
\vspace{-300mm}

\clearpage
\bibliographystyle{splncs04}
\bibliography{main.bib}

\end{document}